\documentclass[runningheads,a4paper]{llncs}

\usepackage{times}
\usepackage{amsmath,amssymb}
\usepackage{graphicx}
\usepackage{url}
\usepackage{todonotes}

\newcommand{\keywords}[1]{\par\addvspace\baselineskip
\noindent\keywordname\enspace\ignorespaces#1}

\renewcommand{\paragraph}[1]{\textbf{#1.}}

\urldef{\mailsa}\path|{sebastian.bayerl,korbinian.riedhammer}@th-nuernberg.de|

\begin{document}

\title{A Comparison of Hybrid and End-to-End Models for Syllable Recognition}

\titlerunning{A Comparison of Hybrid and End-to-End Models for Syllable Recognition}

\author{Sebastian P. Bayerl \and Korbinian Riedhammer}

\institute{Technische Hochschule Nürnberg Georg Simon Ohm, \textsc{Germany}\\
\mailsa\\
}

\toctitle{} \tocauthor{}

\maketitle
\begin{abstract}

This paper presents a comparison of a traditional hybrid speech recognition system (kaldi using WFST and TDNN with lattice-free MMI) and a lexicon-free end-to-end (TensorFlow implementation of multi-layer LSTM with CTC training) models for German syllable recognition on the Verbmobil corpus.
The results show that explicitly modeling prior knowledge is still valuable in building recognition systems.
With a strong language model (LM) based on syllables, the structured approach significantly outperforms the end-to-end model.
The best word error rate (WER) regarding syllables was achieved using kaldi with a 4-gram LM, modeling all syllables observed in the training set.
It achieved 10.0\% WER w.r.t. the syllables, compared to the end-to-end approach where the best WER was 27.53\%.
The work presented here has implications for building future recognition systems that operate independent of a large vocabulary, as typically used in a tasks such as recognition of syllabic or agglutinative languages, out-of-vocabulary techniques, keyword search indexing and medical speech processing.

\end{abstract}

\section{Introduction}\label{sec:intro}

Modeling syllables instead of words is a frequent choice for languages of syllabic nature.
However, modeling syllables instead of words can also improve tasks related to automatic speech recognition (ASR) such as keyword search or identifying out-of-vocabulary (OOV) words \cite{riedhammer_tagalog_2013,smit2017ism}.
Agglutinative languages, such as Turkish, Finish or Swahili, which form words by putting long successions of word units together can benefit from good syllable recognition as they regularly build new words from basic morphemes \cite[p. 293]{enzyklopaedie_der_sprache}.
Furthermore, speech therapy often relies on assessing different granularities of speech from phrases, words, syllables down to phones.
For example, fluency in speech is often measured in terms of (dis-)fluent syllables or their durations \cite{speech_efficiency_Amir2018,yaruss1997clinical,jani2013procedures,percentageSyllables_2006}.

Traditional word-based ASR systems typically depend on prior knowledge in form of pronunciation dictionaries.
While these can grow up to millions of words, the resulting systems still regularly face OOV events.
Phoneme recognition could help to avoid those, however error rates are typically around 20-35\% and thus not reliable enough \cite{lopes2011phoneme}.
This forms the motivation for syllable recognition.
A languages such as German has about 40 phonemes and, due to its feature to form compound words, a near infinite vocabulary.
Although it is hard to come up with a definite number, datasets suggest that about 3000 syllables -- most of them of rare count -- are sufficient to model large vocabularies.\footnote{The vocabulary of the later used Verbmobil data consists 2825 distinct syllables.}
The average length of German words are 1.83 syllables \cite[p. 87]{enzyklopaedie_der_sprache}.

End-to-End models have become increasingly popular in speech recognition over the last couple of years.
Their success suggests that modeling prior knowledge of natural language has become insignificant.
Especially Baidu's \emph{Deep Speech} implementation has taken great part in establishing end-to-end speech recognition as a viable alternative to the traditional hybrid systems that combine hidden Markov models (HMM) and deep neural networks (DNN) \cite{hannun2014first,deepSpeech2}.
These systems work under the assumption, that they are able to learn the acoustic model (AM) and the language model (LM) in a combined effort without the need for specialized prior-infused systems for each task.
Results on large datasets indicate that no explicit language model is needed and prior knowledge about the language is in fact not important to successfully create ASR systems \cite{graves_ctc_2006,deepSpeech2,sak2015learning,DBLP:journals/corr/HannunCCCDEPSSCN14}.
Due to the complexity of the models and the vast amount of data necessary, end-to-end systems require substantial resources to train which makes it relatively expensive to adapt these systems to new requirements, such as a different vocabulary or domain.
A huge advantage of the traditionally structured systems is that they can in part be trained very quickly and easily adapted to a new context or new requirements.
Another drawback of most end-to-end approaches is that they lack accurate time alignment information, which is needed for applications such as keyword search or paralinguistic speech processing.

\paragraph{Our Contribution}
The contribution of this paper is a detailed comparison of a traditional hybrid HMM/DNN system and end-to-end multi-layer long short-term memory (LSTM) network trained using connectionist temporal classification (CTC) for syllable-based ASR using a medium-sized corpus of spontaneous German speech.

\section{Data}\label{sec:data}
For all experiments conducted for this paper, the Verbmobil (VM) corpus was used which comprises of recordings of the \emph{VM1} and \emph{VM2} partitions.
It contains recordings of spontaneous dialogs of people making appointments and general travel planning over the phone.
All systems were trained and evaluated on the published training and evaluation partitions.\footnote{Dataset definitions can be obtained from \url{bas.uni-muenchen.de}}
The dataset used consists of 24435 utterances in 962 dialogs spoken by 629 speakers, the provided lexicon contains 9036 German words.
The training set amounts to about 46 hours of audio.
All audio data was sampled at 16 kHz and is exclusively in German \cite{wahlster2013}.
We chose the Verbmobil data over the larger Voxforge corpus since written text, and thus its read speech, is much more structured and fluent than spontaneous speech -- which is the typical data for the applications described above.
Since the LSTMs will jointly learn the language model, we preferred to stick to data which is closer related to the applications in question.

The lexicon with hand-labeled syllable boundary information was provided by the Friedrich Alexander University (FAU).
It is a result of merging lexica that were created for the research projects EVAR \cite{niemann1985evar}, Verbmobil \cite{wahlster2013}, SmartKom \cite{Reithinger2003smartkom} and SmartWeb \cite{wahlster2004smartweb}.
Guidelines used in its creation closely match the ones described in \cite{batliner2001}.
The lexicon itself was incomplete since it contained only words from the VM1 partition of the dataset.
The missing 2963 entries in the lexicon were generated by using the phonetisaurus grapheme-to-phoneme (g2p) model trained on the FAU syllable lexicon.\footnote{G2P tool available online at \url{https://github.com/AdolfVonKleist/Phonetisaurus}}
For about 40 words, the g2p model was unable to produce a proper syllable representation and those were thus manually generated.  
The syllable transliteration of the training text were obtained by translating the text using the previously generated German syllable lexicon for the VM corpus.

\section{Method}\label{sec:method}

The experiments in this paper were conducted using the Kaldi \cite{kaldi2011} and SRILM toolkits \cite{srilm_2012}.
The end-to-end systems were trained using the multi-purpose machine learning toolkit TensorFlow \cite{tensorflow2015-whitepaper}, using the same Kaldi-generated features.
We use two ``views'' on the data:
Six experiments were conducted using all (2825) syllables and six experiments were conducted using a subset of 199 syllables.
This was done to reduce the number of targets for the CTC training and to reduce model size of the hybrid system.
This split may seem arbitrary at first but was made because 80\% of the dataset consisted of the most common 199 syllables.
The remainder of the syllables amounted to only 20\% of the data.
Even though structured models handle many syllables well, it is interesting to study how that number affects their performance and to do a full fair comparision of the two modeling approaches.

The features used for both systems are 40-dimensional Mel Frequency Cepstral Coefficients (MFCC ``hi-res'') with cepstral mean and variance normalization (CMVN) applied.
Although i-vectors result in slightly better performance \cite{Peddinti2015aspire}, we did not include those since the point of our experiments is to compare traditional hybrid and end-to-end models.

\subsection{Hybrid System}

The setup of the kaldi speech recognition system is based on the kaldi Wall Street Journal (WSJ) recipe, which is a hybrid HMM/DNN system.
For the training of the AM, the ``nnet3 chain'' implementation was used, which uses a time-delay neural network (TDNN) architecture with maximum mutual information (MMI) sequence-level objective function \cite{tddn_2015,povey2016chain_mmi}.
The targets for the DNN training are the context-dependent states obtained by forced alignments of an HMM/GMM baseline system.

For the reduced syllable set, the less common syllables were removed from the lexicon and therefore regarded as unknown (unk).
Additionally, silence, laughter, noise, vocalized noise and unintelligible sounds were added to the lexicon.
The basis for the experiments was a common word-based speech recognition system trained on the VM dataset yielding a word error rate (WER) of 8.7\% with a lexicon containing 9036 words using a 4-gram word based LM with a perplexity of 59.53.
Based on this system, the lexicon was replaced by the previously generated syllable lexicon, and the language model was replaced by a syllable-based language model.

\begin{table}[]
		\caption {Syllable language model perplexities} \label{tab:perplexities}
	\begin{center}
		\begin{tabular}{|l|l|l|}
			\hline
			& Reduced set & Full set \\ \hline
			0-gram           &      201           &        2835      \\ \hline
			1-gram           &      88.97             &      287.59       \\ \hline
			4-gram &         21.49    &  19.97   \\ \hline
		\end{tabular}
	\end{center}

\end{table}

\subsection{End-to-End System}

Long short-term memory (LSTM) units are a special kind of unit in recurrent neural networks (RNN) that are able to learn long-term dependencies through a specific gating mechanism.
The connectionist temporal classification (CTC) loss function uses a blank symbol and an enumeration of all possible alignments.
This makes it slower but independent of per-frame time alignments and allows training deep neural networks for tasks that require long context \cite{graves_ctc_2006}.
For the end-to-end training, the original word-based transcripts were mapped to syllable transliterations.

\begin{figure}[htb]
	\begin{minipage}[b]{1.0\linewidth}
		\centering
		\centerline{\includegraphics[width=8.5cm]{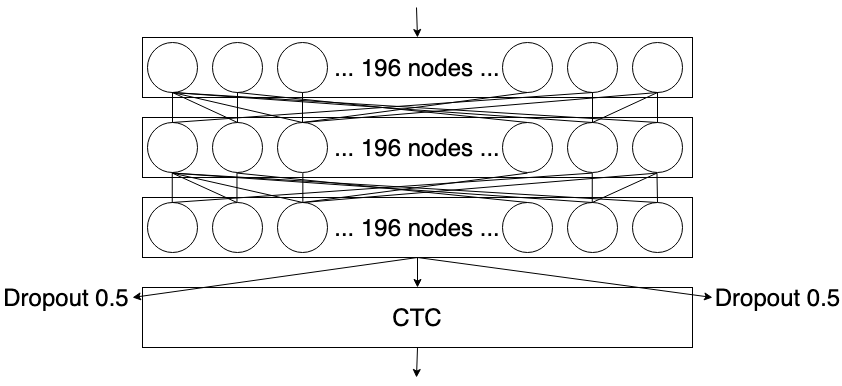}}
	\end{minipage}
	\caption{Topology of the CTC-trained bLSTM network}
	\label{fig:topology}
\end{figure}

The basic network type used in this paper were bidirectional LSTMs.
This is especially useful in situations when future events can help to disambiguate current events.
They were chosen because they produce good results for ASR \cite{graves_hybrid_bidir_2013,graves_speechRNN_LSTM_BI_2013}.
The script used for training the models was based on Ford deepDSP's \emph{deepSpeech} implementation \footnote{base implementation source code available online under \url{https://github.com/fordDeepDSP/deepSpeech}} which is based on Deep Speech 2 \cite{deepSpeech2}.
All end-to-end models were trained using an ADAM optimizer. 
Learning rate decay was used to adapt the learning rate as training progressed from 0.001 down to 0.00003.
The experiments were performed with three different network topologies.
For the first two end-to-end experiments, a network with a single hidden layer and 256 nodes, was used.
The following two experiments were conducted with three fully connected hidden layers of 196 nodes each.
For the final two experiments, we added regularization in form of dropout at training time (50\%).
Otherwise, the network still consisted of 3 hidden layers with 196 nodes each.
Fig.~\ref{fig:topology} shows the final topology.
The output of the network was taken as is and no post-processing applied.
At the time of writing, no lattice generation or LM-based prefix beam search was performed with the output of the bLSTM networks.

\section{Experiments}\label{sec:experiments}

\subsection{Hybrid System}

The baseline models to be compared to the end-to-end models that are assumed to learn AM an LM in one training, 0-gram (no prior) and 1-gram (weak prior) LMs were trained.
The average length of German words is about 1.83 syllables \cite[p. 87]{enzyklopaedie_der_sprache}, thus a 4-gram syllable LM represents contextual knowledge about the composition of words plus some context.
Perplexities for the language models evaluated on the test set can be found in Tab.~\ref{tab:perplexities}.
The LMs for all experiments always used the complete training set with either a complete or a reduced syllable set.

The results for experiments performed with kaldi can be found on the left part of Tab.~\ref{tab:combined}.
As expected, the experiments with the 0-gram language model show very weak performance since they foremost rely on the AM.
This confirms the important role of the LM in hybrid systems.
The WER\footnote{For the remainder of this article, WER is computed w.r.t. the syllable transcription.} for models with more syllables is consistently better; likewise, higher order LMs consistently perform better.
With a WER of 10\% the model with the 4-gram LM and the complete lexicon yields the best result overall.
There is a huge improvement in WER from the 0-gram to the 1-gram model.
Even a very simple LM with no contextual information leads to this huge improvements.
With fewer syllables, the WER still improves but also shows problems of LM that rely on learned sequences and lose some of their power when facing lots of unknown words.
The test dataset consists of about 37,000 syllable instances of which about 6,800 instances were regarded as OOV (about 18\%).
This helps to put results for the reduced syllable experiments into perspective since decoding performance suffers significantly from a high OOV rate.

\subsection{End-2-End System}

Tab.~\ref{tab:combined} shows the WER results for the CTC-trained bLSTM experiments using three different network topologies.
The results indicate that more layers lead to better performance, with a best WER result of 27.53\% using the reduced number of syllables, dropout for regularization and three hidden layers.
This confirms the assumption that deeper networks perform better and the network does better with fewer training targets.
The difference in performance between reduced and full syllable sets almost vanishes for the networks where no dropout was performed.
However, the reduction of targets does not appear to be a deciding factor:
Relative improvement is only about 7.5\% in WER and does not appear reasonable compared to the amount of information lost.
On the other hand, adding dropout and more layers leads to a 24\% relative improvement in performance for the experiments with the complete target set.

\begin{table*}[t]

\caption{WER results for experiments with the hybrid system compared to the end-to-end systems}\label{tab:combined}
\centering
	\begin{tabular}{|l||c|c|c||c|c|c|}
		\hline
		& \multicolumn{3}{c||}{\textbf{Hybrid}} & \multicolumn{3}{c|}{\textbf{End-to-End}}     \\ \hline
		System & 0-gram    & 1-gram    & 4-gram           & 1 Layer & 3 Layer & 3 Layer + dropout \\ \hline\hline
		Reduced Set & 44.6      & 43.0      & 32.9             & 37.22   & 33.89   & \textbf{27.53}    \\ \hline
		Full Set & 58.7      & 35.2      & \textbf{10.0}    & 39.36   & 33.96   & 29.76             \\ \hline
	\end{tabular}
\end{table*}

\section{Discussion}

A surprising observation is that the number of training targets (ie. syllables) does not seem to have a strong impact on the WER results as previously assumed for the bLSTM training.
With more layers, the difference between decoding results almost vanishes just by adding additional layers.
Dropout improves performance up to 18\%, showing the importance of regularization and also confirming the assumption that depth is important \cite{graves_speechRNN_LSTM_BI_2013,deepSpeech2}.
The bLSTM results are a bit worse than expected which could be due to the relatively small dataset.

The hybrid ASR system generally performs better with more targets as soon as it has some useful context information through the LM.
This shows the huge impact of the LM on the decoding results.
As expected, the results for the decoding with more unknown syllables are much worse than with all targets and a 4-gram LM.
Also, the flexibility of the hybrid model is to be mentioned.
Training times for the bLSTM networks were extensive, especially when compared to the swiftness of LM training and decoding graph generation.
The hybrid system can easily be adapted to fit a new lexicon and language model.
The results for this medium-sized dataset were significantly better.

Even though syllables and words are units of different size and error rates are not directly comparable without performing post-processing on the syllable recognition results, it is still interesting to see that error rates for syllable recognition are only slightly worse than the results for word recognition on this dataset.
For future work, it might be important to evaluate the accuracy of the time alignments which are important to many downstream tasks such as keyword search or paralinguistic analyses.

\section{Conclusion and Outlook}\label{sec:outlook}

The direct training of an end-to-end system for syllable recognition is feasible but not as reliable as the hybrid system using a LM, which may be different with more training data.
For this specific task with on a medium-sized corpus, the hybrid approach yields significantly better results.
To achieve better performance with the bLSTMs, its output needs to be combined with LM based prefix beam search, or to train the syllable network along with a LM as proposed in \cite{deepSpeech3_coldfusion_2017}.

For future experiments, we plan to investigate syllable-to-word transduction in order to build an end-to-end system that has high word accuracy while requiring little memory and CPU usage.

\label{sec:refs}

\bibliographystyle{splncs03}

\bibliography{refs}
\end{document}